\documentclass[letterpaper, 10 pt, conference]{ieeeconf}  
\usepackage[utf8]{inputenc}

\IEEEoverridecommandlockouts                              

\usepackage{times}
\usepackage{epsfig}
\usepackage{graphicx}
\usepackage{cite}
\makeatletter
\let\NAT@parse\undefined
\makeatother
\usepackage{amsmath,amssymb,amsfonts}
\usepackage{import}
\usepackage{textcomp}
\usepackage{threeparttable}
\usepackage{xcolor}
\usepackage{subfig}
\usepackage{colortbl}
\usepackage{tikz}
\usepackage[noend]{algpseudocode}
\usepackage[linesnumbered,ruled]{algorithm2e}
\usepackage{balance}
\usepackage{url}
\usepackage{multirow}
\usepackage{placeins}
\usepackage[hidelinks]{hyperref}

\graphicspath{{fig/}}

\title{
\LARGE \bf LIVE-GS: Online LiDAR–Inertial–Visual State Estimation \\
and Globally Consistent Mapping with 3D Gaussian Splatting
}

\author{
        Jaeseok Park$^{1,2}$, Chanoh Park$^{1}$, Minsu Kim$^{2}$, Minkyoung Kim$^{3, *}$ and Soohwan Kim$^{2,*}$
        \thanks{
                This work was carried out with the support of RovifyLab, conducted by the Research Grant of Kwangwoon University in 2025, and partly supported by the Institute of Information \& Communications Technology Planning \& Evaluation(IITP)-ITRC(Information Technology Research Center) grant funded by the Korea government(MSIT)(IITP-2026-2024-00437102).
        }
        \thanks{*corresponding author.}
        \thanks{
                $^1$  Jaeseok Park and Chanoh Park are with the RovifyLab E-mails: {\tt\footnotesize \emph{jaeseok.park, chanoh.park}@rovifylab.com}
        }
        \thanks{
                $^2$ Jaeseok Park and Minsu Kim are with Department of AI Applications, and Soohwan Kim is with Department of Information Convergence, Kwangwoon University, Seoul, Republic of Korea. E-mails: {\tt\footnotesize \emph{jaeseok.park, minsukim, kimsoohwan}@kwu.ac.kr}
        }
        \thanks{
                $^3$Minkyoung Kim is with Department of Big Data Analytics, College of Management, Kyung Hee University, Seoul, Republic of Korea. E-mail: {\tt\footnotesize \emph{minkkim@khu.ac.kr}}
        }
}

\begin{document}

\maketitle
\thispagestyle{empty}
\pagestyle{empty}

\begin{abstract}
	While 3D Gaussian Splatting (3DGS) enabled photorealistic mapping, its integration into SLAM has largely followed traditional camera-centric pipelines. As a result, they inherit well-known weaknesses such as high computational load, failure in texture-poor or illumination-varying environments, and limited operational range, particularly for RGB-D setups. On the other hand, LiDAR emerges as a robust alternative, but its integration with 3DGS introduces new challenges, such as the need for tighter global alignment for photorealistic quality and prolonged optimization times caused by sparse data. To address these challenges, we propose LIVE-GS, an online LiDAR-Inertial Visual SLAM framework that tightly couples 3D Gaussian Splatting with LiDAR-based surfels to ensure high-precision map consistency through global geometric optimization. Particularly, to handle sparse data, our system employs a depth-invariant Gaussian initialization strategy for efficient representation and a bounded sigmoid constraint to prevent uncontrolled Gaussian growth. Experiments on public and our datasets demonstrate competitive performance in rendering quality and map-building efficiency compared with representative 3DGS SLAM baselines.
\end{abstract}

\section{INTRODUCTION}

3D Gaussian Splatting (3DGS)~\cite{kerbl3DGaussianSplatting2023} has advanced photorealistic scene reconstruction and enabled low-latency rendering for robotics applications in Simultaneous Localization and Mapping (SLAM). However, early efforts to integrate 3DGS into SLAM have predominantly relied on cameras, which require a vast number of views to build dense maps, posing a significant bottleneck for efficient operation. Moreover, these vision-based methods are inherently susceptible to challenging conditions such as poorly-textured regions (e.g., white walls) and varying illumination, which can lead to tracking and mapping failures. While RGB-D cameras mitigate some depth ambiguity, they share the same vulnerabilities and introduce additional limitations including short operational range and unreliable outdoor performance under strong sunlight.

To overcome these constraints, the community is increasingly turning to LiDAR sensors, which robustly capture precise 3D points over long distances (hundreds of meters) in diverse conditions. The integration of LiDAR has shown the potential to reduce reconstruction time from longer optimization runs to shorter ones, expanding the applicability of 3DGS from indoor scenes to large-scale outdoor environments.

However, combining LiDAR with 3DGS introduces new technical challenges. First, many existing LiDAR-based methods do not prioritize loop closure for global consistency. Photorealistic rendering is exceptionally sensitive to pose errors; even with pose-graph optimization, a higher level of geometric alignment is required than what is typical for trajectory correction. Second, in unobserved regions such as the sky, or for distant objects where LiDAR points are sparse, Gaussians can become distorted or grow uncontrollably, potentially occluding properly optimized splats behind them. This can degrade map fidelity.

\begin{figure}[t!]
  \includegraphics[width=\linewidth]{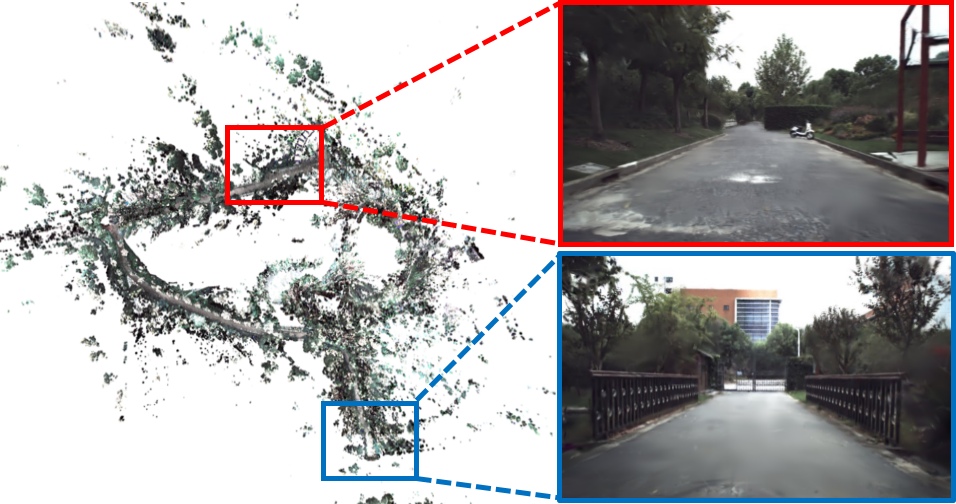}
  \caption{\textbf{LIVE-GS.} Bird's-eye view of a reconstructed 3D Gaussian map (left) and rendered images from selected viewpoints (right). Our globally optimized LiDAR-inertial-visual mapping produces photorealistic renderings in large-scale outdoor environments.}
  \label{fig:teaser}
\end{figure}

\begin{figure*}[!t]
        \includegraphics[width=1\textwidth]{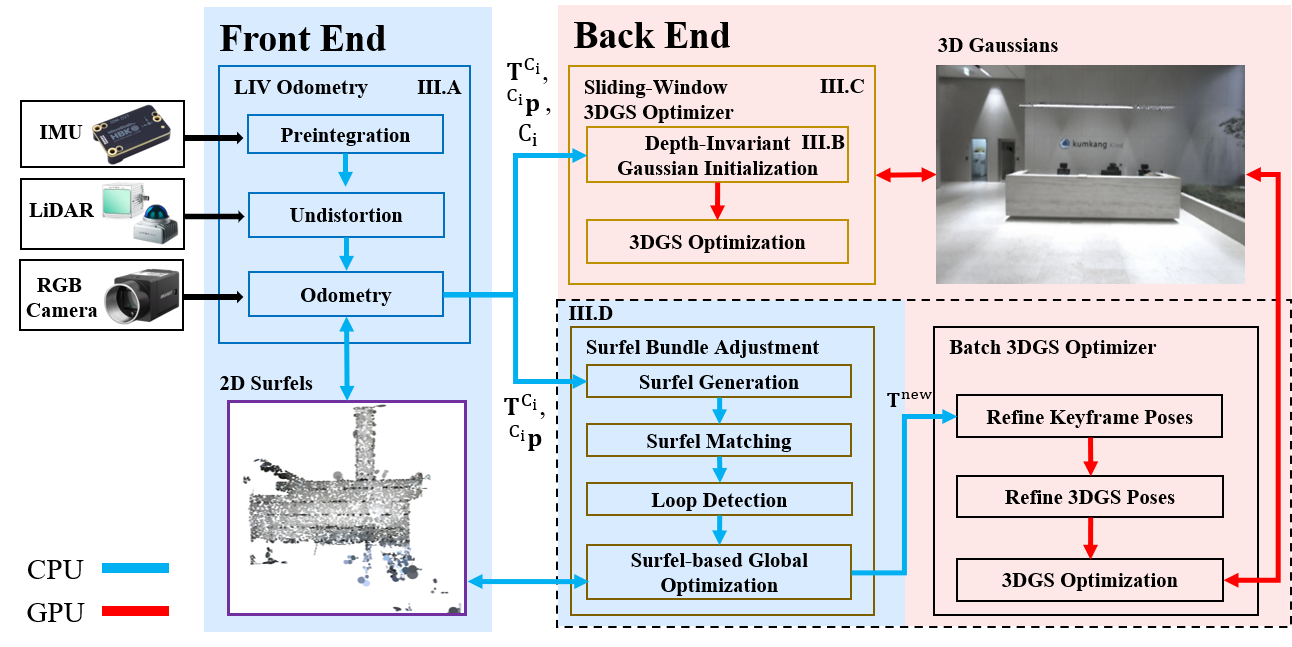}
        \caption{\textbf{LIVE-GS System Overview.}
                Dual-map architecture fuses LiDAR, camera, and IMU streams, tracking poses on a 2D surfel map while optimizing a depth-invariant 3D Gaussian map for photorealistic rendering. Loop closure then performs global refinement that jointly optimizes poses and Gaussians, yielding a geometrically consistent high-resolution map.}
        \label{fig:sov}
\end{figure*}

To address these challenges, we propose LIVE-GS, a novel LiDAR-Inertial-Visual SLAM framework that tightly couples 3D Gaussian mapping with surfel-based optimization. For high-precision alignment, we incorporate a surfel-to-surfel consistency constraint into the pose-graph optimization, which precisely aligns new observations with the existing map structure at the surface level. For efficient distant-object representation, we introduce a depth-invariant initialization strategy that determines Gaussian scale based on its projected footprint in pixel space, rather than by Euclidean distance. Finally, to suppress geometric distortions in unobserved areas, we apply a bounded sigmoid function to constrain the maximum scale of Gaussians.

We focus on improving trajectory consistency and rendering efficiency in large-scale outdoor SLAM with four contributions:
\begin{itemize}
        \item An online LiDAR-Inertial-Visual SLAM framework coupling 3DGS with Surfel Bundle Adjustment (SBA), achieving competitive trajectory accuracy and rendering quality across tested outdoor sequences among online 3DGS methods.
        \item A depth-invariant Gaussian initialization that adapts scale to the projected pixel footprint, reducing map size with only minor PSNR loss.
        \item A bounded sigmoid reparameterization constraining Gaussian scale to a learnable range, suppressing excessive growth in sparsely observed regions.
        \item Comprehensive evaluation on six datasets demonstrating competitive rendering quality and efficiency against both offline and online baselines.
\end{itemize}
\section{RELATED WORK}
\subsection{Offline 3DGS with Pre-computed Poses}
The seminal 3DGS~\cite{kerbl3DGaussianSplatting2023} established an offline paradigm that first computes camera poses via SfM (e.g., COLMAP~\cite{schonbergerStructurefromMotionRevisited2016}) before optimizing scene representation. This paradigm was extended by GS-SDF~\cite{liuGSSDFLiDARAugmentedGaussian2025}, which improved geometric accuracy with LiDAR regularization, and LetsGo~\cite{cuiLetsGoLargeScaleGarage2024}, which utilized LiDAR trajectories for large-scale scenes. However, whether poses come from SfM or advanced LiDAR-Inertial odometry~\cite{shanLVISAMTightlycoupledLidarVisualInertial2021,zhengFASTLIVO2FastDirect2025}, map generation remains a separate offline step. This separation means that early pose errors are fixed before optimization begins, propagating into the final map as ghosting artifacts and reducing suitability for online deployment. These limitations motivated the development of online systems that tightly couple localization and mapping.
\subsection{Online Visual-centric 3DGS}
Subsequent online systems were primarily based on visual SLAM, but their reliance on visual information introduced critical vulnerabilities. For instance, GS-ICP-SLAM~\cite{haRGBDGSICPSLAM2024} relies on a photometric ICP approach, making it prone to tracking failure in texture-poor regions, while Photo-SLAM~\cite{huangPhotoSLAMRealTimeSimultaneous2024} uses feature-based tracking that struggles under abrupt motion or lighting changes. Although RGB-D sensors were introduced in SplaTAM~\cite{keethaSplaTAMSplatTrack2024} to regularize geometry, their short operational range and poor performance under strong sunlight confine them to smaller indoor scenes. Without geometric constraints, these systems also suffer from uncontrolled Gaussian growth in unobserved regions (e.g., sky), where oversized Gaussians occlude those behind them and degrade rendering quality.

\subsection{Online LiDAR-Inertial-Visual 3DGS}
The latest research has focused on combining the robustness of LiDAR with the photorealism of 3DGS, leveraging LiDAR's resilience to challenging lighting and textures as well as its precise long-range depth measurements. However, early hybrids such as Gaussian-LIC~\cite{langGaussianLICRealTimePhotoRealistic2024} retained a vision-centric core and can exhibit trajectory drift and rendering artifacts. More recent methods like LI-GS~\cite{jiangLIGSGaussianSplatting2025} and HGS-Mapping~\cite{wuHGSMappingOnlineDense2024} successfully fuse the sensors but leave two key challenges unaddressed.

These systems rely on sparse pose-graph optimization, which aligns the sensor trajectory at the keyframe level but lacks the dense, surface-level constraints needed for photorealism. Residual pose inconsistencies between the dense Gaussian map and the sparse pose graph in earlier systems, including Gaussian-LIC~\cite{langGaussianLICRealTimePhotoRealistic2024}, can degrade rendering quality.
Moreover, such methods inherit the standard kNN-based initialization from the original 3DGS, which is ill-suited for the non-uniform density of LiDAR data. For sparse, distant points, this strategy creates an excessive number of ill-shaped Gaussians, bloating memory and slowing rendering. Our framework addresses both challenges.
\section{METHODOLOGY}
Our system employs dual map representations: a sparse surfel map for efficient localization and a dense 3D Gaussian map for photorealistic rendering, inspired by Elastic LiDAR Fusion~\cite{parkElasticLiDARFusion2018,parkElasticityMeetsContinuousTime2022}. 

As shown in Fig.~\ref{fig:sov}, the front-end processes synchronized LiDAR-camera-IMU streams for low-latency tracking against the global surfel map while initializing and locally refining Gaussians. Keyframes are created when relative translation or rotation exceeds a threshold. The back-end maintains both maps, performs loop closure with surfel-to-surfel constraints, and globally optimizes poses and Gaussians. Updated maps feed back to the front-end, ensuring long-term consistency.



\subsection{LiDAR-Inertial-Visual Odometry}
We estimate the relative pose between consecutive keyframes by jointly optimizing geometric and photometric constraints. The process begins with propagating the system state (pose, velocity, and bias) from keyframe $i$ to $j$ using IMU preintegration~\cite{forsterOnManifoldPreintegrationRealTime2017}, which provides the initial pose estimate for the subsequent optimization.

For the geometric constraint, we extract surfels from the LiDAR point cloud of each keyframe and establish correspondences across adjacent frames~\cite{parkProbabilisticSurfelFusion2017}. A point-to-plane ICP residual is then formulated for each matched surfel pair $(\mathbf{p}_i, \mathbf{p}_j)$ with its corresponding surface normal $\mathbf{n}_j$:
\begin{equation}
    e_{\text{icp}} = \mathbf{n}_{j}^{\top}(^{W}\mathbf{R}_j^{-1}({^{W}\mathbf{R}_{i}} \mathbf{p}_{i} + {^{W}\mathbf{t}_{i}}- {^{W}\mathbf{t}_{j}}) - \mathbf{p}_j) \,,
    \label{eq:point-to-plane_ICP}
\end{equation}
where $(^{W}\mathbf{R}_i, ^{W}\mathbf{t}_i)$ and $(^{W}\mathbf{R}_j, ^{W}\mathbf{t}_j)$ are the global poses of keyframes $i$ and $j$.

In parallel, for the photometric constraint, we select sparse feature points from the image and track them across keyframes~\cite{forsterSVOFastSemidirect2014}. The photometric residual is defined by directly comparing pixel intensities:
\begin{equation}
    e_{\text{photo}} = \mathbf{I}_{i}(\mathbf{u,v}) - \mathbf{I}_{j}(\pi(^{j}\mathbf{R}_{i} \mathbf{p}_{i} + ^{j}\mathbf{t}_{i}))
    \label{eq:photometric_error}
    \,,
\end{equation}
\begin{equation}
\pi(\mathbf{p})=
\begin{bmatrix}
f_x \frac{X}{Z} + c_x \\
f_y \frac{Y}{Z} + c_y
\end{bmatrix}
\,,
\label{eq:projection_matrix}
\end{equation}
where $\mathbf{I}_{i}(\mathbf{u,v})$ is the intensity at pixel $(\mathbf{u,v})$ in frame $i$, $\mathbf{p}_{i}$ is a 3D LiDAR point in the local frame of keyframe $i$, and $({^j\mathbf{R}_i}, {^j\mathbf{t}_i})$ is the relative transformation from frame $i$ to $j$. The function $\pi(\cdot)$ projects the transformed point onto frame $j$'s image plane, where intensity is bilinearly interpolated.

The full odometry cost function combines these two residuals:
\begin{equation}
{L}_{\text{odom}} = \sum\|e_{\text{icp}}\|^{2} + \lambda_{\text{photo}}\sum\|e_{\text{photo}}\|^{2}
\label{eq:full_odom_error}
\,,
\end{equation}
where $\lambda_{\text{photo}}=0.5$ down-weights the photometric term whose intensity-scale residuals are typically larger than metric-scale ICP residuals. The IMU-propagated pose provides the initial estimate; no explicit IMU cost term is needed. The refined relative pose serves as an odometry constraint in the global pose graph.

\begin{figure}[t]
\centering
\subfloat[]{\def\svgwidth{0.25\textwidth}
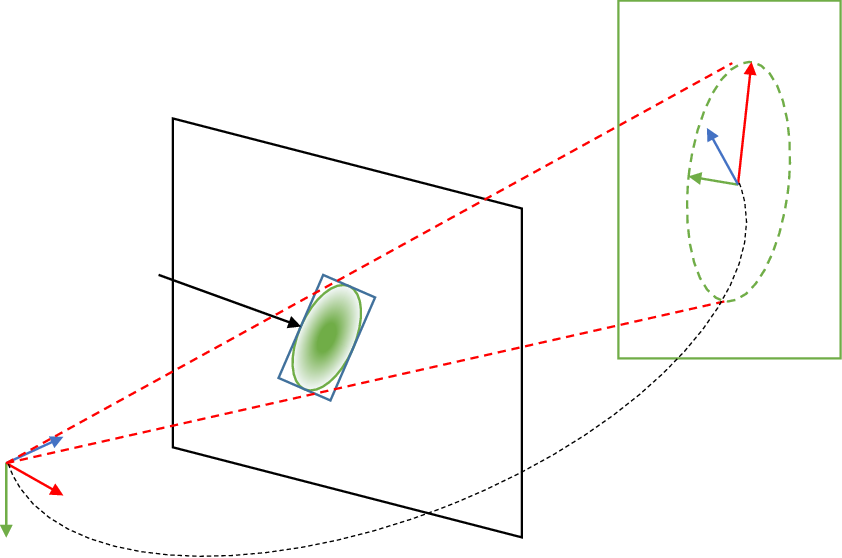}
\subfloat[]{\def\svgwidth{0.14\textwidth}
\begingroup%
  \makeatletter%
  \providecommand\color[2][]{%
    \errmessage{(Inkscape) Color is used for the text in Inkscape, but the package 'color.sty' is not loaded}%
    \renewcommand\color[2][]{}%
  }%
  \providecommand\transparent[1]{%
    \errmessage{(Inkscape) Transparency is used (non-zero) for the text in Inkscape, but the package 'transparent.sty' is not loaded}%
    \renewcommand\transparent[1]{}%
  }%
  \providecommand\rotatebox[2]{#2}%
  \newcommand*\fsize{\dimexpr\f@size pt\relax}%
  \newcommand*\lineheight[1]{\fontsize{\fsize}{#1\fsize}\selectfont}%
  \ifx\svgwidth\undefined%
    \setlength{\unitlength}{168.80120087bp}%
    \ifx\svgscale\undefined%
      \relax%
    \else%
      \setlength{\unitlength}{\unitlength * \real{\svgscale}}%
    \fi%
  \else%
    \setlength{\unitlength}{\svgwidth}%
  \fi%
  \global\let\svgwidth\undefined%
  \global\let\svgscale\undefined%
  \makeatother%
  \begin{picture}(1,1.30797432)%
    \lineheight{1}%
    \setlength\tabcolsep{0pt}%
    \put(0,0){\includegraphics[width=\unitlength]{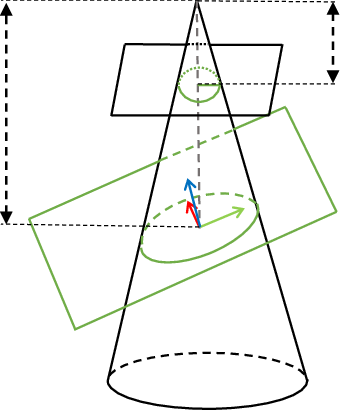}}%
    \put(0.73451525,0.42981301){\color[rgb]{0,0,0}\makebox(0,0)[lt]{\lineheight{1.25}\smash{\begin{tabular}[t]{l}\fontsize{9pt}{1em} ${\bold\Sigma_{3D}}$\end{tabular}}}}%
    \put(0.243970469,1.07801094){\color[rgb]{0,0,0}\makebox(0,0)[lt]{\lineheight{1.25}\smash{\begin{tabular}[t]{l}\fontsize{9pt}{1em} ${I}$\end{tabular}}}}%
    \put(-0.01007241,0.8091337){\color[rgb]{0,0,0}\makebox(0,0)[lt]{\lineheight{1.25}\smash{\begin{tabular}[t]{l}\fontsize{9pt}{1em} ${Z}$\end{tabular}}}}%
    \put(0.6376531,1.0923199){\color[rgb]{0,0,0}\makebox(0,0)[lt]{\lineheight{1.25}\smash{\begin{tabular}[t]{l}\fontsize{9pt}{1em} ${\bold\Sigma_{2D}}$\end{tabular}}}}%
    \put(0.9576531,1.05923199){\color[rgb]{0,0,0}\makebox(0,0)[lt]{\lineheight{1.25}\smash{\begin{tabular}[t]{l}\fontsize{5pt}{1em} ${f}$\end{tabular}}}}%
  \end{picture}%
\endgroup%
}
\caption{
    \textbf{Depth-Invariant Gaussian Initialization.}  (a) illustrates the overall process of depth-invariant Gaussian initialization, while (b) shows its relationship with conics
}
\label{fig:pai}
\end{figure}

\subsection{Depth-Invariant Gaussian Initialization}
Conventional LiDAR-based 3D Gaussian Splatting (3DGS) methods initialize Gaussian scales from local point density. This approach has a significant drawback: it creates overly large and blurry Gaussians in sparse and distant regions. To address this, our Depth-Invariant Gaussian Initialization strategy initializes 3D scales to ensure their projection onto the image plane maintains a consistent and predefined size, regardless of depth or density.

For each LiDAR point $\mathbf{p}$, we first compute the covariance matrix via a k-Nearest Neighbors (KNN) search and Principal Component Analysis (PCA) to obtain a rotation matrix $\mathbf{R}$. We then initialize the Gaussian with anisotropic scales $\sigma_{x}^{2}$ and $\sigma_{y}^{2}$ in the tangent plane and scales $\sigma_{z}^{2}$ along the normal. The 3D covariance matrix is thus defined as:
\begin{equation}
\label{eq:scale2cov3}
\boldsymbol\Sigma_{3D} =
\mathbf{R}
\text{diag}(\sigma_{x}^{2}, \sigma_{y}^{2}, \sigma_{z}^{2})
\mathbf{R}^{\top}\,.
\end{equation}

To compute the projected covariance, we define the Jacobian \(\mathbf{J} \in \mathbb{R}^{2 \times 3}\) of the camera projection in Eq.~(\ref{eq:projection_matrix}):
\begin{equation}
\mathbf{J} =
\begin{bmatrix}
\frac{f_x}{Z} & 0 & -\frac{f_x X}{Z^2} \\
0 & \frac{f_y}{Z} & -\frac{f_y Y}{Z^2}
\end{bmatrix}
\label{eq:Jacobian}
\,.
\end{equation}

 As depth $Z$ increases, the Jacobian entries decrease, yielding smaller 2D projections (Fig.~\ref{fig:pai}), which motivates conic-based scale adaptation. The 2D covariance of the projected Gaussian is:
\begin{equation}
\boldsymbol\Sigma_{2D} = \mathbf{J}  \boldsymbol\Sigma_{3D}  \mathbf{J}^\top
\,.
\label{eq:cov3d2cov2d}
\end{equation}

We compute the area ratio ${k}$ between target area ${A}_\text{target}$ and original projected area ${A}_\text{origin}$, and scale the initial 3D scales by $\sqrt{k}$. Here $A_\text{target} = \pi n^2/4$ for a circular footprint of $n$~pixels in diameter; we evaluate $n{=}1$ and $n{=}5$ in Section~\ref{sec:experiments}:
\begin{equation}
{A}_\text{origin} = \pi \sqrt{\det({\boldsymbol \Sigma_{2D}})},\;\;k=\frac{A_\text{target}}{A_\text{origin}}
\,.
\end{equation}

Since $\Sigma \propto \sigma^2$, scaling each $\sigma$ by $\sqrt{k}$ scales the projected area by $k$:
\begin{equation}
\therefore \sigma_x' = \sigma_x \sqrt{k},\;\;  \sigma_y' = \sigma_y \sqrt{k}, \;\; \sigma_z' = \sigma_z \sqrt{k}
\,.
\end{equation}

This produces Gaussians with adaptive in-plane size, enabling stable rendering across depths.

\begin{algorithm}[!t]
\SetAlgoLined
\KwIn{{Current Gaussian scalings $\{\mathbf{s}_{\text{old}}\}_{i=1}^N$, current thresholds $\sigma_{\text{max}}, \sigma_{\text{min}}$}}
\KwOut{{Updated threshold $\sigma_{\text{max}}$}, $\mathbf{s}_{\text{new}}$}
$\text{max}\sigma \leftarrow \{ \| \text{BoundedSigmoid}(\mathbf{s}_{\text{old}}, \sigma_{\text{max}}) \| \}_{i=1}^N$

$\text{ratio}_{\text{top}} \leftarrow \frac{1}{N} \sum_{i=1}^{N} \left[ \text{max}\sigma_i > 0.95\sigma_{\text{max}} \right]$

$\text{ratio}_{\text{low}} \leftarrow \frac{1}{N} \sum_{i=1}^{N} \left[ \text{max}\sigma_i < 0.05\sigma_{\text{max}} \right]$

\uIf{$\text{ratio}_{\text{top}} > 0.15$}{
    $\text{old} \leftarrow \sigma_{\text{max}}$ \\
    $\mathbf{s}_{\text{new}} \leftarrow \text{BoundedSigmoid}(\mathbf{s}_{\text{old}}, \sigma_{\text{max}}$)
    $\sigma_{\text{max}} \leftarrow 1.2 \cdot \sigma_{\text{max}}$ \\
}
\ElseIf{$\text{ratio}_{\text{low}} > 0.95$}{
    $\text{old} \leftarrow \sigma_{\text{max}}$ \\
    $\mathbf{s}_{\text{new}} \leftarrow \text{InverseBoundedSigmoid}(\mathbf{s}_{\text{old}}, \sigma_{\text{max}}$)
    $\sigma_{\text{max}} \leftarrow \max(0.8 \cdot \sigma_{\text{max}},\ 4 \cdot \sigma_{\text{min}})$ \\
}

\caption{\textbf{Adaptive scaling adjustment of Gaussian splats.} The upper threshold $\sigma_{\text{max}}$ is increased when too many splats are saturated, and decreased when most are small, maintaining a balanced distribution of splat sizes.}
\label{alg:adapt_sigma_max}
\end{algorithm}

\subsection{Local GS Mapping}\label{sec:local_mapping}
\begin{figure}[!t]
\centering
\def\svgwidth{70mm}
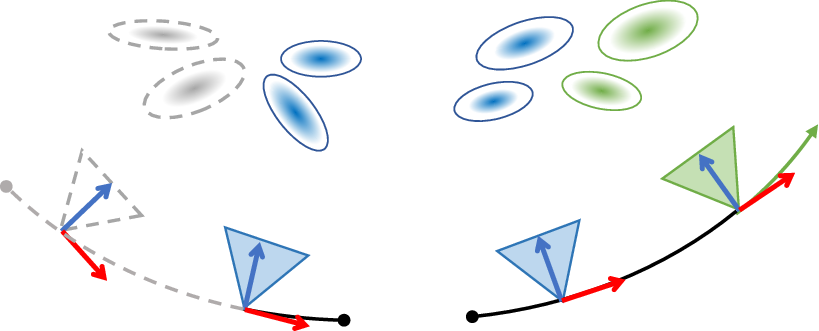
\caption{
\textbf{Sliding-window 3DGS Optimization.}
At index~$i$, we jointly optimize the most recent $K$ camera keyframes
$\{\mathbf{T}_{i-K+1},\dots,\mathbf{T}_{i}\}$ together with all Gaussians visible from them.
}
\label{fig:sliding_window}
\end{figure}

While the initially assigned parameters of each Gaussian capture the rough structure of the environment based on LiDAR data, further refinement is required to achieve precise alignment and photorealistic rendering. As illustrated in Fig.~\ref{fig:sliding_window}, we refine the map by optimizing the parameters of visible Gaussians within a fixed-size sliding window $W=\{I_{i-K+1},\dots,I_i\}$ that contains the most recent $K$ keyframes.

The rendered pixel \(\hat{I}(u,v)\) is computed using an alpha compositing equation that accumulates view-dependent colors weighted by opacity~\cite{kerbl3DGaussianSplatting2023}:
\begin{equation}
\hat{I}(u,v) = \sum_{g} T_g \cdot \alpha_g \cdot C_g(\mathbf{v}_g), \quad T_g = \prod_{k=1}^{g-1} (1 - \alpha_k)
\label{eq:alphablending}
\,,
\end{equation}
where Gaussians are sorted in front-to-back order based on their projected depth.

To quantify the discrepancy between the rendered image and the observed image, we define a rendering loss that combines the L1 loss and the Structural Dissimilarity Index (D-SSIM) loss~\cite{wangImageQualityAssessment2004}:
\begin{equation}
{L}_{\text{render}} = (1 - \lambda_{\text{ssim}}){L}_{\text{L}1} + \lambda_{\text{ssim}}{L}_{\text{D-SSIM}}
\label{eq:rendering_loss}
\,,
\end{equation}
where $\lambda_{\text{ssim}}=0.2$ balances geometric and perceptual similarity.

In regions with insufficient observations such as the sky or background, Gaussians tend to grow excessively large during optimization. To prevent this, we apply a bounded sigmoid reparameterization on the learnable scale parameters ${s}$, mapping them to a bounded range $[\sigma_{\min},\sigma_{\max}]$:
\begin{equation}
\sigma_{\text{bounded}} = \sigma_{\min} + (\sigma_{\max} - \sigma_{\min}) \cdot \text{sigmoid}(s)
\label{eq:boundedsigmoid}
\,.
\end{equation}

The inverse mapping is given by:
\begin{equation}
s = \text{sigmoid}^{-1} \left(\frac{\sigma_{\text{bounded}} - \sigma_{\min}}{\sigma_{\max} - \sigma_{\min}}\right)
\label{eq:inverseboundedsigmoid}
\,.
\end{equation}

While using a fixed upper bound $\sigma_{\text{max}}$ is effective, we further improve stability by dynamically adjusting it based on the current distribution of Gaussian sizes. This adaptive process is detailed in Algorithm~\ref{alg:adapt_sigma_max}, where the BoundedSigmoid and InverseBoundedSigmoid functions correspond to Eq.~(\ref{eq:boundedsigmoid}) and Eq.~(\ref{eq:inverseboundedsigmoid}), respectively.

For high-quality scene reconstruction, we employ a sequential optimization strategy. This process is divided into two distinct steps. First, in the local pose optimization stage, we refine the current camera poses $\{\mathbf{T}_{i-K+1},...,\mathbf{T}_{i}\}$ by minimizing the geometric loss ${e}_{\text{icp}}$. Subsequently, with the pose held fixed, the Gaussian refinement stage optimizes the parameters of visible Gaussians by minimizing the photometric rendering loss ${L}_{\text{render}}$.

This two-step optimization process is performed within a sliding window framework to maintain computational efficiency. When a new pose $\mathbf{T}_{i+1}$ is added, the window slides forward by marginalizing the oldest pose $\mathbf{T}_{i-K}$ and its associated Gaussians to maintain a fixed problem size. This approach decouples pose estimation from map refinement while improving trajectory consistency and rendering quality over time.

\subsection{Global GS Mapping}
While local mapping improves geometric alignment and photorealistic rendering within short temporal windows, it does not explicitly address long-term drift or inconsistencies that arise during large-scale exploration. To mitigate these issues, we perform global optimization over the full trajectory using a pose-graph framework that incorporates both motion and geometric constraints.

Each keyframe in the SLAM trajectory is represented as a node in the global pose graph, and two types of edges connect them, which are relative odometry constraints derived from the local front-end and surfel-to-surfel geometric alignments between spatially or temporally distant keyframes. The former ensures continuity, while the latter introduces long-range structure-preserving terms~\cite{parkElasticLiDARFusion2018,parkElasticityMeetsContinuousTime2022}.

The surfel-based alignment error between two keyframes \(i\) and \(j\) is formulated using the G-ICP~\cite{segalGeneralizedICP2009} framework. Correspondences are found by nearest-neighbor search; pairs exceeding a Mahalanobis distance threshold are rejected as outliers. The residual vector is:
\begin{equation}
\mathbf{e}_{\text{surfel}}(i, j) = {}^i \mathbf{R}_j \mathbf{p}_j + {}^i \mathbf{t}_j - \mathbf{p}_i
\label{eq:global_surfel_con}
\,,
\end{equation}
with the combined covariance of the matched surfel pair:
\begin{equation}
\mathbf{C}_{ij} = \boldsymbol{\Sigma}_{i}+ {}^{i}\mathbf{R}_{j}\boldsymbol{\Sigma}_{j} {}^{i}\mathbf{R}_{j}^\top
\label{eq:combined_cov}
\,,
\end{equation}
where $\boldsymbol{\Sigma}_i$ and $\boldsymbol{\Sigma}_j$ are the covariances of matched surfels in their respective local frames, and ${}^{i}\mathbf{R}_{j}$ transforms $\boldsymbol{\Sigma}_j$ from frame $j$ into frame $i$ following the G-ICP formulation~\cite{segalGeneralizedICP2009}.

Global optimization minimizes the sum of both relative pose errors and surfel-based residuals over all pairs of connected keyframes:
\begin{equation}
{L}_{\text{global}} \hspace{-0.2em}=\hspace{-0.2em} \sum_{i, j} \| e_{\text{pose}}(i, j) \|^2_{\boldsymbol\Sigma_{\text{odom}}} \hspace{-0.2em}+\hspace{-0.2em} \lambda_{\text{surfel}} \sum_{i, j} \| \mathbf{e}_{\text{surfel}}(i, j) \|^2_{\mathbf{C}_{ij}}
\label{eq:pose_only_BA}
\,,
\end{equation}
where $\|\mathbf{e}\|^2_{\boldsymbol\Sigma} = \mathbf{e}^\top \boldsymbol\Sigma^{-1} \mathbf{e}$ denotes the squared Mahalanobis norm. $e_{\text{pose}}(i,j)$ is the relative pose constraint from odometry with noise covariance $\boldsymbol\Sigma_{\text{odom}}$, and $\lambda_{\text{surfel}}=1.0$ balances geometric alignment with motion consistency.

To solve this optimization efficiently and incrementally, we employ iSAM2~\cite{kaessISAM2IncrementalSmoothing2012}, which updates only the affected parts of the pose graph without relinearizing the full system at every iteration. This allows our method to scale to long-term operations while maintaining computational efficiency.

After the global pose-graph optimization converges, only a subset of keyframes \(\mathbf{T}^{\text{refined}}\) are typically affected. We update the Gaussians associated with these refined keyframes to reflect their new poses. Specifically, for each keyframe $\mathbf{T} \in \mathbf{T}^{\text{refined}}$, we apply the following transformation to its associated Gaussians ${G}^{\text{refined}}$:
\begin{equation}
\forall \mathbf{G} \in {G}^{\text{refined}} ,
\quad
\mathbf{p}_{{G}}^{\text{new}} = \mathbf{R}_{\mathbf{T}}^{\text{opt}} \mathbf{p}_{G}^{\text{old}} + \mathbf{t}_{\mathbf{T}}^{\text{opt}}
\label{eq:refine_translate}
\,,
\end{equation}

\begin{algorithm}[!t]
\SetAlgoLined

\KwIn{Local Keyframe Trajectory $\mathbf{T}^{\text{kf}}$}
\KwOut{Refined Global Trajectory $\hat{\mathbf{T}}^{\text{kf}}$}

\text{optimized, loop} $\leftarrow$ $\text{GetTrajectory}(\mathbf{T}^{\text{kf}})$

\If{optimized}{
        \ForEach{$\mathbf{T}_i^{\text{kf}} \in \mathbf{T}^{\text{kf}}$ not in $\hat{\mathbf{T}}_i^{\text{kf}}$}{
            $\hat{\mathbf{T}}_i^{\text{kf}} \leftarrow \hat{\mathbf{T}}_{i-1}^{\text{kf}} \cdot \left(\mathbf{T}_{i-1}^{\text{kf}}\right)^{-1} \mathbf{T}_i^{\text{kf}}$}
        Apply trajectory refinement using $\hat{\mathbf{T}}^{\text{kf}}$

        \If{loop closure detected}{
            Trigger sparse map refinement \\
}
}
\caption{\textbf{Surfel Bundle Adjustment Keyframe Refinement.} Periodically retrieves optimized global keyframe trajectory $\hat{\mathbf{T}}^{\text{kf}}$ from the SBA. If loop closure is detected, a sparse map refinement process is triggered.}
\label{alg:refine_thread}
\end{algorithm}

\begin{equation}
\mathbf{R}_{G}^{\text{new}} = \mathbf{R}_{\mathbf{T}}^{\text{opt}} \mathbf{R}_{G}^{\text{old}}
\label{eq:refine_rotate}
\,,
\end{equation}
where \(\mathbf{p}_{{G}}^{\text{old}}, \mathbf{R}_{{G}}^{\text{old}}\) are the original Gaussian position and orientation, and \(\mathbf{R}_{\mathbf{T}}^{\text{opt}}, \mathbf{t}_{\mathbf{T}}^{\text{opt}}\) are the optimized keyframe pose.

Following this transformation, the spatial configuration of the Gaussians is aligned with the globally optimized keyframe trajectory. However, since these pose updates may introduce inconsistencies in appearance or surface coverage, we perform a second-stage refinement of the updated Gaussians to restore photorealistic rendering quality.

This refinement focuses on internal Gaussian parameters such as scale, opacity, and spherical harmonics coefficients while keeping the camera poses fixed. We reuse the rendering loss function defined in Section~\ref{sec:local_mapping}, restricting its application to the updated Gaussian batch ${G}^{\text{refined}}$:
\begin{equation}
{L}^{\text{refined}}_{\text{render}} = \sum_{\mathbf{T} \in \mathbf{T}^{\text{refined}}} {L}_{\text{render}}
\label{eq:refine_render_loss}
\,,
\end{equation}
where ${L}_{\text{render}}^{\text{refined}}$ denotes the rendering loss for a single keyframe, using the previously defined combination of L1 and SSIM terms.

This refinement is performed using the Adam optimizer and ensures that the appearance of the Gaussians remains consistent with observations, even after global pose adjustments. By limiting optimization to the affected regions, we preserve both visual fidelity and computational efficiency.

\section{EXPERIMENTS}
\label{sec:experiments}

\subsection{Experimental Setup}
We evaluate public and self-collected datasets spanning large-scale outdoor and small-scale indoor environments. From the FAST-LIVO2 dataset~\cite{zhengFASTLIVO2FastDirect2025}, we use three urban sequences: Retail\_Street (Retail), CBD02, and SYSU\_01 (SYSU01). We also adopt the UAV-based airport03 sequence from MARS-LVIG~\cite{liMARSLVIGDatasetMultisensor2024}, which covers challenging coastal terrain. Furthermore, we use three sequences (00, 01, 07) from the Botanic Garden dataset~\cite{linR3LIVERobustRealTime2022} for trajectory accuracy evaluation, as it provides centimeter-level RTK-GPS ground truth. 
For indoor evaluation, we collected two custom datasets, Reception and Demo room, in office-like and exhibition-style environments with constrained geometry and limited texture.

In Tables~\ref{tbl:comparison_result_left} and~\ref{tbl:comparison_result_right}, we compare six datasets against GS-SDF~\cite{liuGSSDFLiDARAugmentedGaussian2025}, LetsGo~\cite{cuiLetsGoLargeScaleGarage2024}, Photo-SLAM~\cite{huangPhotoSLAMRealTimeSimultaneous2024}, GS-ICP-SLAM~\cite{haRGBDGSICPSLAM2024}, and two variants of our method (Ours(1pixel), Ours(5pixel)) using PSNR, map size, and total duration. The ablation study in Table~\ref{tbl:ablation_study} additionally uses the Reception sequence from our custom collection.

\begin{figure}[!t]
	\centering
	\includegraphics[width=1\linewidth]{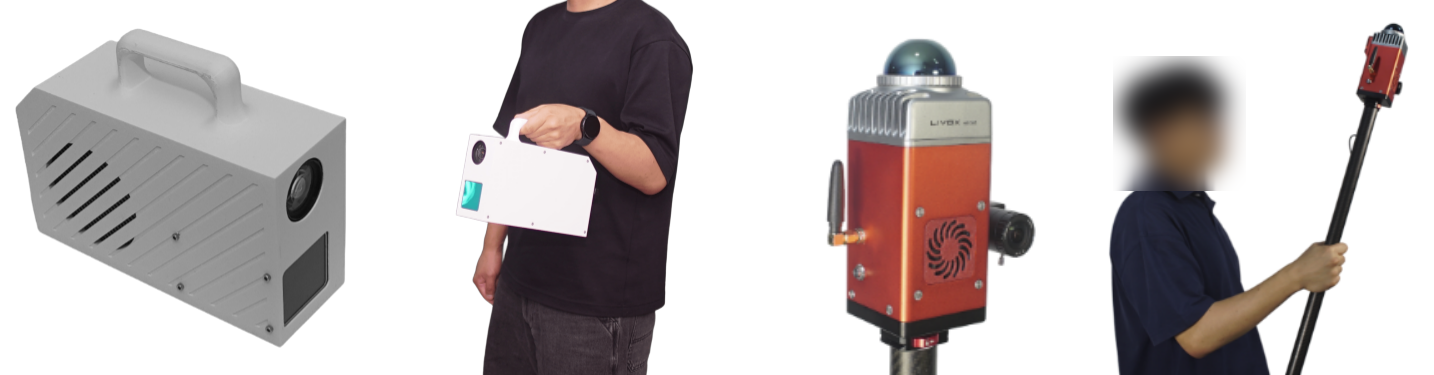}
	\caption{\textbf{Experimental handheld platforms.} The Livox AVIA (left) and MID-360 (right) are integrated with IMU and camera sensors.}
	\label{fig:exps}
\end{figure}

For our custom datasets, camera intrinsics and LiDAR-Camera extrinsics were calibrated using the motion-based toolkit~\cite{parkSpatiotemporalCameraLiDARCalibration2020}, and precise temporal alignment was ensured by our custom handheld device with hardware-level synchronization among LiDAR, camera, and IMU, as shown in Fig.~\ref{fig:exps}.
All experiments were conducted on a desktop with an AMD Ryzen 9 7900 CPU, 64\,GB RAM, and an NVIDIA RTX 4090 GPU. For depth-invariant initialization, we report two variants: Ours(1pixel) and Ours(5pixel), corresponding to target projected footprints of 1 and 5 pixels, respectively.

\subsection{Trajectory Accuracy Evaluation}

We evaluate localization accuracy on the Botanic Garden dataset, which provides centimeter-level RTK-GPS ground truth. We compare against two categories of baselines: traditional LIV methods (FAST-LIVO2~\cite{zhengFASTLIVO2FastDirect2025}, R3LIVE~\cite{linR3LIVERobustRealTime2022}) and LIV-based 3DGS methods (GS-LIVM~\cite{jianGSLIVMRealTimeLargeScale2025}, Gaussian-LIC~\cite{langGaussianLICRealTimePhotoRealistic2024}).

Among 3DGS-based methods, GS-LIVM~\cite{jianGSLIVMRealTimeLargeScale2025} exhibits competitive ATE on Seq.~01 (0.557\,m) but shows larger errors on Seq.~07 (1.785\,m), whereas Gaussian-LIC~\cite{langGaussianLICRealTimePhotoRealistic2024} shows consistently higher errors across all sequences. This suggests that naive integration of 3DGS with LiDAR-inertial odometry is insufficient for maintaining long-term trajectory consistency. In contrast, our surfel bundle adjustment enforces surfel-to-surfel consistency constraints across keyframes, helping suppress accumulated drift in extended sequences.

\begin{figure}[!t]
        \centering
        \includegraphics[width=\columnwidth]{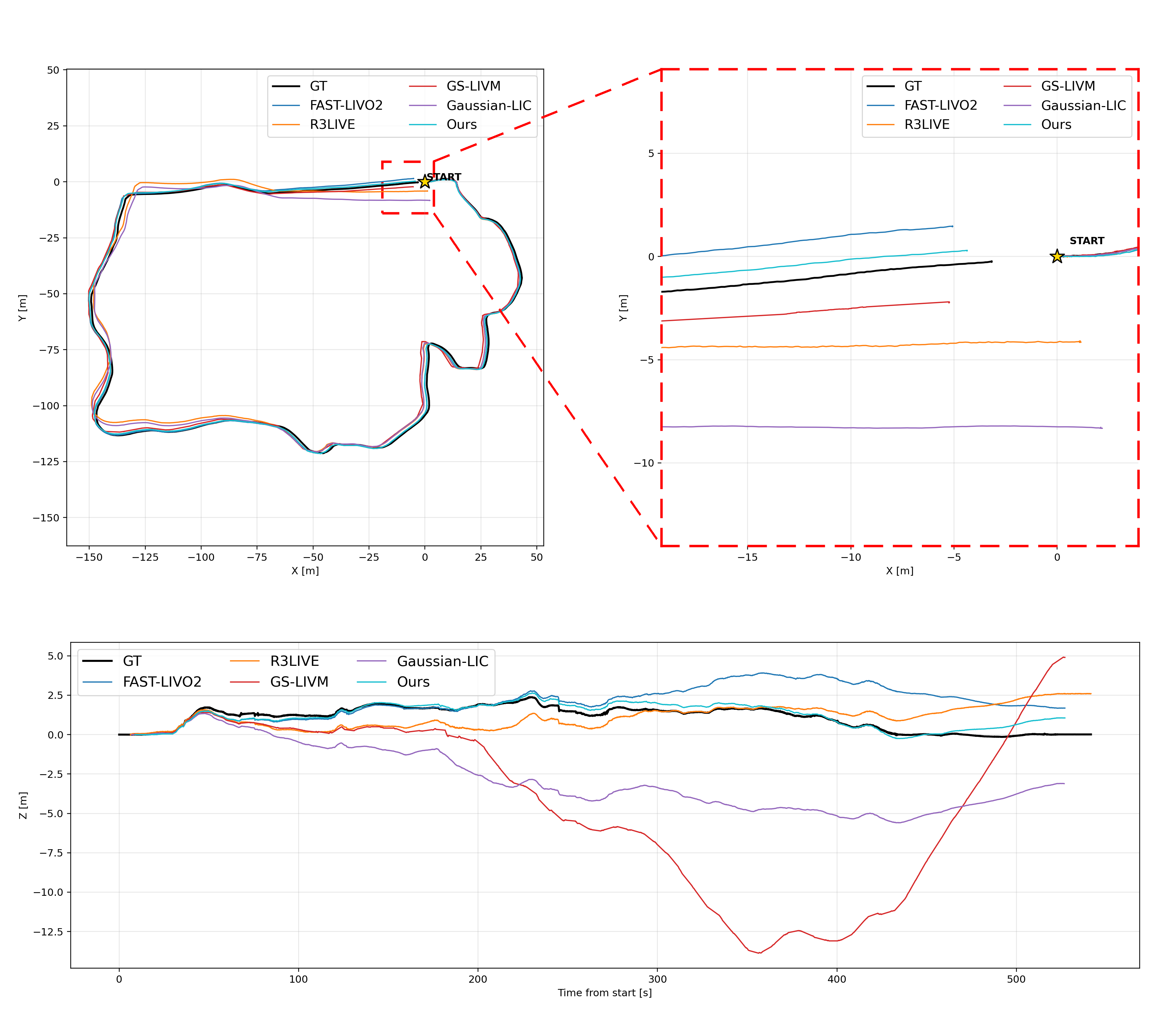}
        \caption{\textbf{Trajectory accuracy on Botanic Garden.} Our method achieves the closest alignment to the RTK-GPS ground-truth trajectory on Seq. 07.}
        \label{fig:trajectory_accuracy}
\end{figure}

As shown in Table~\ref{tbl:trajectory_accuracy} and Fig.~\ref{fig:trajectory_accuracy}, Ours achieves the best ATE on all three sequences (0.528/0.529/0.498\,m). For Z-RMSE, Ours is best on Seq.~00/01 (0.352/0.177\,m), while FAST-LIVO2~\cite{zhengFASTLIVO2FastDirect2025} is best on Seq.~07 (0.281\,m) and Ours is second (0.286\,m). FAST-LIVO2's advantage in vertical accuracy on Seq.~07 likely stems from its direct LiDAR feature alignment that tightly constrains vertical drift; nevertheless, our method achieves the best overall ATE on the same sequence (0.498\,m vs.\ 0.784\,m), indicating improved 3D localization on the reported sequences.

\subsection{Rendering Comparison with Existing Methods}
Tables~\ref{tbl:comparison_result_left} and~\ref{tbl:comparison_result_right} compare our LIVE-GS with offline methods GS-SDF~\cite{liuGSSDFLiDARAugmentedGaussian2025} (30k iterations) and LetsGo~\cite{cuiLetsGoLargeScaleGarage2024} (15k iterations), as well as online methods Photo-SLAM~\cite{huangPhotoSLAMRealTimeSimultaneous2024} and GS-ICP-SLAM~\cite{haRGBDGSICPSLAM2024}, in terms of reconstruction fidelity (PSNR), final map size (Map), and full-sequence processing time (Dur.).

\begin{table}[!t]
        \centering
        \caption{\textbf{Trajectory accuracy on the Botanic Garden dataset~\cite{linR3LIVERobustRealTime2022}.}}
        \begin{threeparttable}
        \begin{tabular}{lllcc}
                \hline\noalign{\smallskip}
                Dataset & Category                   & Method            & ATE (m)$\downarrow$   & Z-RMSE (m)$\downarrow$ \\
                \hline\noalign{\smallskip}
                \multirow{5}{*}{Seq.~00}
                        & \multirow{2}{*}{Trad.}
                        & FAST-LIVO2                 & 1.190             & 0.677                                  \\
                        &                            & R3LIVE            & \cellcolor{orange!15}0.825 & \cellcolor{orange!15}0.422  \\
                                \cline{2-3}
                        & \multirow{3}{*}{3DGS}
                        & GS-LIVM                    & 1.247             & 0.498                                  \\
                        &                            & Gaussian-LIC      & 1.812             & 0.698              \\
                        &                            & Ours              & \cellcolor{green!25}\textbf{0.528}    & \cellcolor{green!25}\textbf{0.352}     \\
                \hline\noalign{\smallskip}
                \multirow{5}{*}{Seq.~01}
                        & \multirow{2}{*}{Trad.}
                        & FAST-LIVO2                 & 0.731             & 0.239                                  \\
                        &                            & R3LIVE            & 0.583             & 0.199              \\
                                \cline{2-3}
                        & \multirow{3}{*}{3DGS}
                        & GS-LIVM                    & \cellcolor{orange!15}0.557 & \cellcolor{orange!15}0.179                      \\
                        &                            & Gaussian-LIC      & 0.783             & 0.410              \\
                        &                            & Ours              & \cellcolor{green!25}\textbf{0.529}    & \cellcolor{green!25}\textbf{0.177}     \\
                \hline\noalign{\smallskip}
                \multirow{5}{*}{Seq.~07}
                        & \multirow{2}{*}{Trad.}
                        & FAST-LIVO2                 & \cellcolor{orange!15}0.784 & \cellcolor{green!25}\textbf{0.281}                         \\
                        &                            & R3LIVE            & 1.986             & 0.673              \\
                                \cline{2-3}
                        & \multirow{3}{*}{3DGS}
                        & GS-LIVM                    & 1.785             & 1.575                                  \\
                        &                            & Gaussian-LIC      & 2.832             & 0.728              \\
                        &                            & Ours              & \cellcolor{green!25}\textbf{0.498}    & \cellcolor{orange!15}0.286  \\
                \hline\noalign{\smallskip}
        \end{tabular}
        \begin{tablenotes}
                \footnotesize
                \item \textit{Note:} \colorbox{green!25}{\textbf{Green}} and \colorbox{orange!15}{Orange} denote the best and second-best results, respectively.
        \end{tablenotes}
        \end{threeparttable}
        \label{tbl:trajectory_accuracy}
\end{table}

\begin{figure*}[!t]
        \centering
        \includegraphics[width=\linewidth]{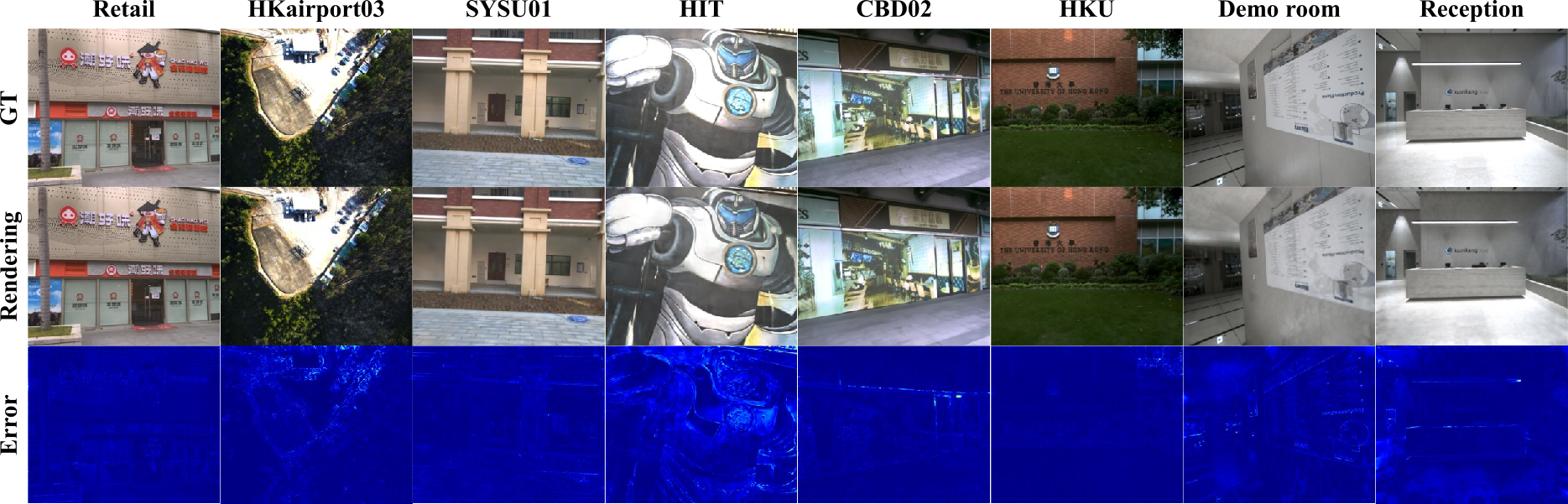}
        \caption{\textbf{Qualitative rendering results.} Each column is a dataset; rows show GT, rendered image, and error map against GT}
        \label{fig:render_result}
\end{figure*}

Our Ours(1pixel) variant achieves the highest PSNR on the CBD02, Retail, and airport03 datasets among the compared baselines. The corresponding qualitative results are presented in Fig.~\ref{fig:render_result}. The Ours(5pixel) variant reduces the map size with only a minor PSNR decrease, achieving the smallest map on five of six datasets (e.g., Retail 31.0\,MB, SYSU01 24.7\,MB). On our in-house datasets, our method records the fastest processing time (96\,s) with the smallest map (26.8\,MB). Notably, on Demo room, Ours(5pixel) surpasses Ours(1pixel) in PSNR (28.3 vs.\ 25.5\,dB) because close-range capture yields sparser point clouds relative to the scene extent; the larger initial pixel footprint of Ours(5pixel) better covers this sparsity, producing more complete Gaussians. Thus, our method achieves comparable rendering quality while keeping both time and storage lower than offline approaches.

\begin{table}[!t]
        \centering
        \caption{\textbf{Rendering comparison with existing methods on the FAST-LIVO2~\cite{zhengFASTLIVO2FastDirect2025} datasets.}}
        \begin{threeparttable}
                \footnotesize
                \setlength{\tabcolsep}{3pt}
                \renewcommand{\arraystretch}{0.9}
                \begin{tabular}{llrrr}
                        \hline
                        Dataset & Method       & PSNR$\uparrow$   & Map(MB)$\downarrow$ & Dur.(s)$\downarrow$ \\
                        \hline
                        \multirow{6}{*}{CBD02}
                                & GS-SDF       & 24.1             & 287.1               & 589                 \\
                                & LetsGo       & 24.3             & 203.0               & 516                 \\
                        \cline{2-2}
                                & Photo-SLAM   & 20.2             & \cellcolor{orange!15}171.7   & \cellcolor{orange!15}319     \\
                                & GS-ICP-SLAM  & 18.1             & 288.3               & 321                 \\
                                & Ours(1pixel) & \cellcolor{green!25}\textbf{27.5}    & 235.0               & \cellcolor{green!25}\textbf{296}        \\
                                & Ours(5pixel) & \cellcolor{orange!15}27.0 & \cellcolor{green!25}\textbf{55.2}       & \cellcolor{green!25}\textbf{296}        \\
                        \hline
                        \multirow{6}{*}{Retail}
                                & GS-SDF       & 27.2             & 451.1               & 1032                \\
                                & LetsGo       & \cellcolor{orange!15}29.0 & 293.2               & 343                 \\
                        \cline{2-2}
                                & Photo-SLAM   & 28.3             & 84.6                & 153                 \\
                                & GS-ICP-SLAM  & 18.7             & \cellcolor{orange!15}74.3    & \cellcolor{orange!15}147     \\
                                & Ours(1pixel) & \cellcolor{green!25}\textbf{29.2}    & 111.7               & \cellcolor{green!25}\textbf{145}        \\
                                & Ours(5pixel) & 28.1             & \cellcolor{green!25}\textbf{31.0}       & \cellcolor{green!25}\textbf{145}        \\
                        \hline
                        \multirow{6}{*}{SYSU01}
                                & GS-SDF       & 25.6             & 308.0               & 1752                \\
                                & LetsGo       & 26.7             & 307.5               & 486                 \\
                        \cline{2-2}
                                & Photo-SLAM   & 23.6             & \cellcolor{orange!15}81.3    & \cellcolor{orange!15}193     \\
                                & GS-ICP-SLAM  & 18.3             & 275.2               & 399                 \\
                                & Ours(1pixel) & \cellcolor{orange!15}27.8 & 90.2                & \cellcolor{green!25}\textbf{183}        \\
                                & Ours(5pixel) & \cellcolor{green!25}\textbf{27.9}    & \cellcolor{green!25}\textbf{24.7}       & \cellcolor{green!25}\textbf{183}        \\
                        \hline
                \end{tabular}
                \begin{tablenotes}
                        \footnotesize
                        \item \textit{Note:} \colorbox{green!25}{\textbf{Green}} and \colorbox{orange!15}{Orange} denote the best and second-best results, respectively. - indicates algorithm failure.
                \end{tablenotes}
        \end{threeparttable}
        \label{tbl:comparison_result_left}
\end{table}

\begin{figure}[!t]
	\centering
	\subfloat{\includegraphics[width=0.52\linewidth]{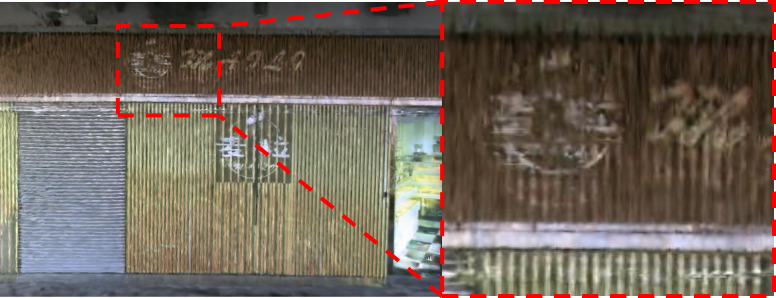}}%
	\subfloat{\includegraphics[width=0.48\linewidth]{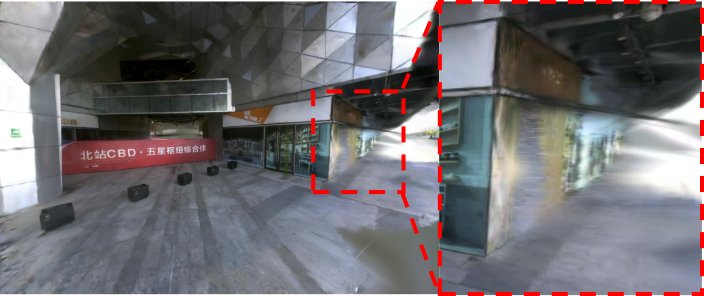}}%
	\\[-1em]
	\subfloat{\includegraphics[width=0.52\linewidth]{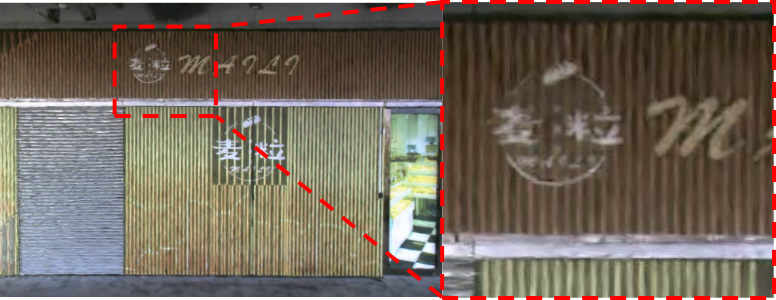}}%
	\subfloat{\includegraphics[width=0.48\linewidth]{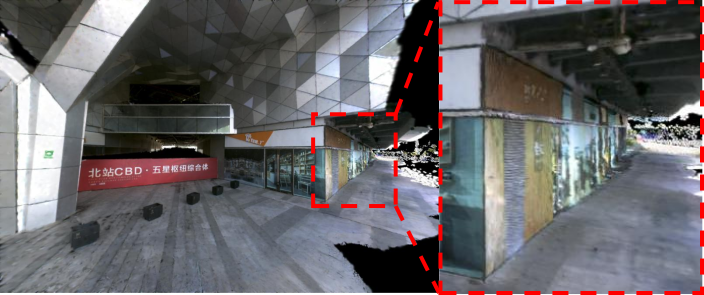}}
	\caption{\textbf{Qualitative ablation results of BS and SBA.} The top row shows results without SBA on the left and without BS on the right. The bottom row displays results with SBA on the left and with BS on the right.}
	\label{fig:abl}
\end{figure}

\begin{table}[!t]
        \centering
        \caption{\textbf{Rendering comparison with existing methods on the MARS-LVIG~\cite{liMARSLVIGDatasetMultisensor2024} and our datasets.}}
        \begin{threeparttable}
                \footnotesize
                \setlength{\tabcolsep}{3pt}
                \renewcommand{\arraystretch}{0.9}
                \begin{tabular}{llrrr}
                        \hline
                        Dataset & Method       & PSNR$\uparrow$   & Map(MB)$\downarrow$ & Dur.(s)$\downarrow$ \\
                        \hline
                        \multirow{6}{*}{airport03}
                                & GS-SDF       & -                & -                   & -                     \\
                                & LetsGo       & 17.3             & 329.1               & 843                   \\
                        \cline{2-2}
                                & Photo-SLAM   & 16.9             & \cellcolor{orange!15}77.9    & \cellcolor{green!25}\textbf{265}          \\
                                & GS-ICP-SLAM  & -                & -                   & -                     \\
                                & Ours(1pixel) & \cellcolor{green!25}\textbf{20.0}    & 146.4               & \cellcolor{orange!15}334       \\
                                & Ours(5pixel) & \cellcolor{orange!15}19.7 & \cellcolor{green!25}\textbf{47.1}       & \cellcolor{orange!15}334       \\
                        \hline
                        \multirow{6}{*}{Demo room}
                                & GS-SDF       & \cellcolor{orange!15}28.1 & 199.9               & 952                   \\
                                & LetsGo       & 25.4             & 107.7               & 277                   \\
                        \cline{2-2}
                                & Photo-SLAM   & \cellcolor{orange!15}28.1 & \cellcolor{green!25}\textbf{17.2}       & 170                   \\
                                & GS-ICP-SLAM  & 22.9             & 87.0                & \cellcolor{green!25}\textbf{46}           \\
                                & Ours(1pixel) & 25.5             & 166.5               & \cellcolor{orange!15}142       \\
                                & Ours(5pixel) & \cellcolor{green!25}\textbf{28.3}    & \cellcolor{orange!15}57.5    & \cellcolor{orange!15}142       \\
                        \hline
                        \multirow{6}{*}{Reception}
                                & GS-SDF       & 26.2             & 314.1               & 1427                  \\
                                & LetsGo       & \cellcolor{green!25}\textbf{27.7}    & 31.5                & 209                   \\
                        \cline{2-2}
                                & Photo-SLAM   & 22.6             & \cellcolor{orange!15}31.4    & \cellcolor{orange!15}117       \\
                                & GS-ICP-SLAM  & 22.7             & 181.6               & 174                   \\
                                & Ours(1pixel) & 23.4             & 94.8                & \cellcolor{green!25}\textbf{96}           \\
                                & Ours(5pixel) & \cellcolor{orange!15}26.7 & \cellcolor{green!25}\textbf{26.8}       & \cellcolor{green!25}\textbf{96}           \\
                        \hline
                \end{tabular}
                \begin{tablenotes}
                        \footnotesize
                        \item \textit{Note:} \colorbox{green!25}{\textbf{Green}} and \colorbox{orange!15}{Orange} denote the best and second-best results, respectively. - indicates algorithm failure.
                \end{tablenotes}
        \end{threeparttable}
        \label{tbl:comparison_result_right}
\end{table}

\begin{table}[!t]
	\centering
	\caption{\textbf{Quantitative ablation results of BS and SBA}}
	\begin{threeparttable}
		\begin{tabular}{llrrrr}
			\hline\noalign{\smallskip}
			        &                                 & BS(x)  & BS(x)         & BS(O)  & BS(O)         \\
			Dataset & Metric                          & SBA(x) & SBA(O)        & SBA(x) & SBA(O)        \\
			\hline\noalign{\smallskip}

			\multirow{2}{*}{CBD02}
                        & PSNR$\uparrow$                  & 25.9   & \cellcolor{orange!15}26.4          & 26.0   & \cellcolor{green!25}\textbf{27.0} \\
                        & \textbf{max} $\sigma\downarrow$ & 690.23 & 599.54        & \cellcolor{orange!15}0.43   & \cellcolor{green!25}\textbf{0.43} \\
			\hline\noalign{\smallskip}

			\multirow{2}{*}{Retail}
                        & PSNR$\uparrow$                  & 27.6   & \cellcolor{orange!15}27.9          & 27.8   & \cellcolor{green!25}\textbf{28.1} \\
                        & \textbf{max} $\sigma\downarrow$ & 71.29  & 59.04         & \cellcolor{orange!15}0.34   & \cellcolor{green!25}\textbf{0.34} \\
			\hline\noalign{\smallskip}

			\multirow{2}{*}{Reception}
                        & PSNR$\uparrow$                  & \cellcolor{orange!15}27.0   & \cellcolor{green!25}\textbf{27.2} & 24.5   & 26.7          \\
                        & \textbf{max} $\sigma\downarrow$ & 105.17 & 154.07        & \cellcolor{orange!15}0.12   & \cellcolor{green!25}\textbf{0.11} \\
			\hline\noalign{\smallskip}
		\end{tabular}
		\begin{tablenotes}
			\footnotesize
			\item \textit{Note:} BS denotes the Bounded Sigmoid, and SBA denotes the Surfel Bundle Adjustment. \colorbox{green!25}{\textbf{Green}} and \colorbox{orange!15}{Orange} denote the best and second-best results, respectively.
		\end{tablenotes}
	\end{threeparttable}
	\label{tbl:ablation_study}
\end{table}

Offline methods incur substantial computational overhead. For example, GS-SDF~\cite{liuGSSDFLiDARAugmentedGaussian2025} requires up to 1752\,s on SYSU01, and LetsGo~\cite{cuiLetsGoLargeScaleGarage2024} consumes 307.5\,MB on the same dataset, whereas online baselines face robustness challenges. GS-ICP-SLAM~\cite{haRGBDGSICPSLAM2024} and GS-SDF fail entirely on airport03 due to algorithm failure in large-scale outdoor environments. Our Ours(5pixel) variant provides a practical balance between quality and efficiency by initializing Gaussians at a coarser projected footprint, reducing map size (e.g., CBD02: 235.0 $\rightarrow$ 55.2\,MB, a 4.3$\times$ reduction) with only a marginal PSNR decrease (27.5 $\rightarrow$ 27.0\,dB). These results suggest that depth-invariant initialization can control representation density while maintaining perceptual fidelity.

\subsection{Ablation Study}

We ablate the two key components of our system. Surfel Bundle Adjustment (SBA) and the Bounded Sigmoid (BS) constraint--to isolate their individual and combined effects. Results are summarized in Table~\ref{tbl:ablation_study} and visualized in Fig.~\ref{fig:abl}.

SBA alone improves PSNR by refining keyframe alignment but does not control Gaussian scale; the maximum scale can remain above the scene extent, producing oversized splats that blur object boundaries (Fig.~\ref{fig:abl}, top-left). BS alone constrains scale to sub-meter levels and restores structural fidelity (Fig.~\ref{fig:abl}, top-right). However, on Reception, which is an indoor scene whose glass walls expose distant outdoor regions, the strict bound removes coarse coverage of far-away areas and lowers PSNR, revealing that scale control alone cannot compensate for pose misalignment in mixed indoor-outdoor settings.

Combining both components yields the best result across all datasets: SBA provides globally consistent poses while BS ensures spatially compact Gaussians, and neither component alone achieves this balance.

\section{CONCLUSION}
We presented LIVE-GS, a LiDAR–Inertial–Visual 3DGS SLAM system with surfel bundle adjustment, depth-invariant initialization, and bounded sigmoid reparameterization. These contributions enhance trajectory accuracy and rendering quality while reducing map size. In the experiments, we demonstrate competitive results across various environments. 

However, our system does not explicitly leverage LiDAR depth supervision during Gaussian optimization. Incorporating depth constraints could enforce tighter geometric consistency and suppress floating Gaussians that arise in textureless or sparsely observed regions. Thus, future work will explore leveraging temporal modeling to handle dynamic objects and incorporating learned priors for appearance decomposition to improve robustness under adverse weather conditions.

{\footnotesize
        \bibliographystyle{IEEEtran}
        \bibliography{ref}
}

\end{document}